\documentclass[twoside,11pt]{article}

\usepackage{jmlr2e}

\usepackage{algorithm}
\usepackage{algpseudocode}
\usepackage{amsmath}
\usepackage{graphicx}
\usepackage{bbm}

\jmlrheading{1}{2019}{1-48}{4/00}{10/00}{Larkin Liu, Richard Downe, \& Joshua Reid}

\ShortHeadings{Multi-Armed Bandit Strategies for Non-Stationarity \& Delayed Feedback}{Liu, Downe, \& Reid 2019}
\firstpageno{1}

\begin{document}

\title{Multi-Armed Bandit Strategies for Non-Stationary Reward Distributions and Delayed Feedback Processes}

\author{\name Larkin Liu \email larkin.liu@loblaw.ca \\
        \name Richard Downe \email rdowne@loblaw.ca \\
       \addr Loblaw Digital \\
       Toronto, ON, Canada
       \AND
       \name Joshua Reid \email js2reid@uwaterloo.ca \\
       \addr University of Waterloo \\
       Waterloo, ON, Canada
}

\maketitle

\begin{abstract}
A survey is performed of various Multi-Armed Bandit (MAB) strategies in order to examine their performance in circumstances exhibiting non-stationary stochastic reward functions in conjunction with delayed feedback. We run several MAB simulations to simulate an online eCommerce platform for grocery pick up, optimizing for product availability. In this work, we evaluate several popular MAB strategies, such as $\epsilon$-greedy, UCB1, and Thompson Sampling. We compare the respective performances of each MAB strategy in the context of regret minimization. We run the analysis in the scenario where the reward function is non-stationary. Furthermore, the process experiences delayed feedback, where the reward function is not immediately responsive to the arm played.  We devise a new Bayesian technique (AG1) tailored for non-stationary reward functions in the delayed feedback scenario. The results of the simulation show show superior performance in the context of regret minimization compared to traditional MAB strategies.\\
\end{abstract}


\section{Introduction}

\subsection{The Multi-Armed Bandit Strategy}

In this paper, we present the formulation, algorithms and practical approaches to the application of different algorithms designed to effectively address the Multi-Armed Bandit (MAB) problem. We further investigate the practical case where the reward distributions of the arms are non-stationary along with the reward having a delayed feedback to the algorithm. This is more practical because in many cases, the time of retrieval of the reward information is not immediate. Furthermore, reward distributions may also change over time (non-stationary) in practical situations due to variance in the subjects being measured to determine the reward of each arm. We present simulated results in this paper to investigate the performance of different MAB strategies for both stationary and non-stationary reward functions. Specifically, we evaluate the performance of the MAB algorithms for the purpose of optimizing inventory availability for online eCommerce in the grocery sector.
 
The application of MAB strategies to the field of optimization has had a wide range of success. Applications of MAB strategies can be found in Web Search \citep{Graepel:2010}, Real-Time-Bidding \citep{Flajolet:2017} and product pricing \citep{Trovo:2015}. In this work, we specifically provide an overview and evaluation of MAB strategies for the purpose of forecasting item availability for online groceries. Each predictive algorithm provides a forecast of item availability for the entire inventory of items, and the MAB strategy will decide the optimal arm to select in order to optimize the empirical probability that each assorted article is actually available at the time of pick-up. It is essential to the operation of the business to be able to quickly identify the best algorithm for product assortment and play the arm the maximizes the metric to reduce the need for customers to substitute or remove items from their shopping list. In the following discourse, we shall study the application of MAB strategies in both the stationary and non-stationary scenario.

\subsection{Industry Overview}

In this paper, we address the issue of item availability via an eCommerce grocery service, created for the Loblaw Corporation, a large grocery store chain in Canada. In their online service, \textit{Click-and-Collect}, customers can order items on the website, and proceed to collect the items in-store, or have the items delivered to their household via a third party. It is vital to the service that the assortment of items displayed on the website is readily available at the time of pick-up. Multiple algorithms run on the inventory and demand data to forecast the availability of these grocery articles. This paper, however, does not explicate the specific details of the in-house forecasting algorithms, rather, it addresses the application of MAB techniques to select the optimal algorithm to forecast the in-store assortment of products. We refer to these algorithms in this text simply as \textit{assortment algorithms}, \textit{models}, or \textit{arms}. There are two major sections of this paper, first we present the theoretical background behind MAB algorithms. Subsequently we present the application of the MAB model in a simulated environment closely mimicking that of an actual online assortment process.

The primary metric, \textit{fill-rate}, of each store is evaluated on a store basis and is the number of items that were in stock in that store when the order was being assembled. In other MAB experiments, multiple metrics are sometimes optimized by a single algorithm, some examples be found in \citep{Oner:2018}. However, in our experiment we will consider the evaluation of each MAB strategy on a single metric, the \textit{fill rate}.

\section{Mathematical Background}
\subsection{Regret Minimization}

We formulate the MAB strategy in the context of regret minimization. We define the Q function for every action $k_t$, as $Q(k_t)$, indicating that the reward obtained from playing arm, $k$, at time step $t$. Provided a policy $\pi$, the expected reward, $V_t$ from taking action $k_t$ can be expressed as,

\begin{equation} \label{eq:Qsa}
	V_t(\pi) = \sum_{}^{k} Q(k_{t}) P(k_{t}| \pi)
\end{equation}

$k_t$ is the action undertaken in the current time, according to policy $\pi$. In our application, we defined $V_t$ as the value of the state at time $t$. However, the reward function, $Q(k_t)$ is unknown. We denote the optimal value at time $t$ as $V^*_t$, and under the optimal policy $\pi^*$. 

\begin{equation} \label{eq:V*}
	V_t(\pi^*) =  \max_{\pi} V_t(\pi)
\end{equation}

The parameters, $\theta$, which generate $Q(k_{t})$ may or may not change with respect to time, depending on the stationarity of the reward function. Provided that any action $k$, can be selected depending on the policy, cumulative regret of a policy, upon completion of the game at the final time epoch $T$, is defined as $R_T$, 

\begin{equation} \label{eq:regret}
	R_T(\pi) = \sum_{t = 0}^{T} \Big[ V_t(\pi^*) - V_t(\pi) \Big]
\end{equation}

Where $Q(k_t)$ is a constant for each $k$ an time interval, $t$, depending on the stipulations in Section \ref{lab:non-stat-bandit}. The MAB strategy's objective is to minimize the cumulative regret from measurements of the reward function of each arm played that epoch.

\subsection{The Non-Stationary Bandit} \label{lab:non-stat-bandit}

We define the expected reward $\mu_{t}^k$ as the expected value of the reward from playing arm $k$ at time $t$, 

\begin{equation} \label{eq:stat-reward}
	 \mu_{t}^k = \mathop{\mathbb{E}}[Q(k_{t})]
\end{equation}

A stationary bandit consists of a reward process where the expected reward of the reward distribution for each arm does not change over time. Hence for the stationary and non-stationary reward function $\mu^k$ and $\mu_{t}^k$ respectively,

\begin{align}
    \mu^k &= f(\theta^k) \label{eq:non-stat-reward} \\
    \mu_{t}^k &= f(t,\theta^k)
\end{align}

Where $\theta^k$ represents the parameters of the expected reward distribution for arm $k$. In the non-stationary case, the reward function becomes a function of $t$ as well as $k$. Where a deterministic function, $f$ controls expected reward $\mu_{t}^k$ for purposes of simulation. Thus the optimal expected reward is calculated as, 

\begin{equation} \label{eq:optimal_expected_reward}
	\mu^{*}_t = \operatorname*{max}_{k \in \{ 1,...,K\} } \mu^k_t 
\end{equation}

For example, \citep{Besbes:2014} investigated the effect of MAB's in the non-stationary case using a sinusoidal function to generate $\mu_{t}^k$, subsequently a confidence bound on the expected regret. The algorithm proposed in \citep{Besbes:2014} was unable to detect changes in the optimal arm dynamically, and relied on constantly resetting their MAB strategy to find the optimal arm in a brief time window. In our work, we propose an algorithm that is capable of detecting the change point of the reward distributions in a short amount of time, and is able to adaptively react to the change point of the expected reward function to further minimize regret.

\subsection{The Delayed Feedback Framework} \label{subsec:delayed_feedback}

In our scenario, we define the concept of \textit{delayed-feedback} as a reward process which does not immediately return an reward value upon playing of a selected arm. This idea is similarly noted in \citep{Chapelle:2011}, where some number of arms need to be played before a reward observation can be obtained from the system. In our case, we specify the number of stores as $N$, and the number of arms, or \textit{assortment algorithms}, as $K$. Each arm is denoted as being the $k^{th}$ arm, where $k \in \{1, ..,K\}$. In our simulation, we specify N = 100, and K = 10. 

In order to simulate our particular scenario, each store, out of N stores, is assigned an arm for the number of items to assort (or to fill). We refer to this number as $\gamma$. We set $\gamma= 50$ for all $N$ stores. We estimate the parameter $\mu^k_t $ of the reward distributions at each delayed feedback interval. We express this estimate as,

\begin{equation} \label{eq:mu_calc}
	\widehat{\mu^k_t} = \frac{1}{Nt} \sum_{t \in \Omega} \sum_{n=1}^N \frac{1}{\gamma} \sum_{i=1}^{\gamma}  \mathbbm{1}(k,\pi) Q(k_t)
\end{equation}

Where $\mathbbm{1}(k,\pi)$ is an indicator function stating if arm $k$ was played under policy $\pi$, representing the MAB strategy. We use $\Omega$ to denote the set of all time epochs, $t$, contained within the observation period. In our delayed feedback scenario, once the store is assigned an arm, $k$, it must play that assortment algorithm for all $\gamma$ items it needs to assort. Inherently, each arm, $k$, has a specified expected reward $\mu^{k}_t$. In the stationary case, $\mu^{k}_t$ is constant and independent of time. In the non-stationary case $\mu^{k}_t$ can vary over time. For some strategies, we compute the parameter estimates for $\mu^{k}_t$ under a renewal framework, where only recent data is used to compute $\widehat{\mu^k_t}$, we refer to this as $\Omega_r$, otherwise all observations from the beginning of the simulation is included in the time window, referred to as $\Omega$,

\begin{align}
    \Omega &= \{1,..., t - 1 \} \label{eq:omega_nr} \\
    \Omega_r &= \{t - r,..., t - 1 \} \label{eq:omega_r}
\end{align}

In the context of delayed feedback, the observations from the previous time epoch $t - 1$ until the end of the renewal period $t - r$ are used to calculate parameter estimates for $\widehat{\mu^k_t}$, as denoted by Eq. (\ref{eq:omega_r}). Else if the MAB strategy is non-renewing, Eq. (\ref{eq:omega_nr}) is used to define the time window.

\section{Multi-Armed Bandit Strategies}

\subsection{Epsilon Greedy Strategy}

In the perspective of total cumulative regret, $R_T$, the $\epsilon$-greedy selection strategy selects the expected optimal arm with probability $1 - \epsilon$, and other potential arm's with probability $p_K$.For n arms, we specify the probability of selecting the other arms as $p_K$ is,

\begin{equation} \label{eq:eg_pn}
	p_K = \frac{\epsilon}{K - 1} 
\end{equation}

In other words, each arm other than the optimal arm selected with probability $p_n$. For example, \citep{Kuleshov:2014} showed that the $\epsilon$-greedy strategy performs significantly better than random experimentation. This is due to the fact the regret is greedily minimized as more observations are collected. However, the regret of the $\epsilon$-greedy algorithm is constant at each iteration, and linear in $t$ due to the fact that other sub-optimal arms will always be played. One can criticize an $\epsilon$-greedy strategy on the grounds that it has poor asymptotic behavior, because it continues to explore long after the optimal solution becomes apparent.

\begin{algorithm}[h!]
\caption{$\epsilon$-Greedy}\label{algo:eg}
\begin{algorithmic}[1]
    \State $Q = \emptyset$
    \For {$t = 0 \to T$}
        \For {$k = 1 \to K$}
            \State Compute $\widehat{\mu^k_t}$ from $\Omega$
        \EndFor
        \State $k = \operatorname*{argmax}_{k} \widehat{\mu^k_t} $
        \State Play $k$ with probability $ 1 - \epsilon$, else play another arm with probability $p_K$ 
        \State $Q_k \gets Q(k) $ 
    \EndFor
\end{algorithmic}
\end{algorithm}

The lower bound on the expected regret of $\epsilon$-greedy is proportional to $T$ linearly in the infinite time horizon as the $\epsilon$-greedy strategy is forced to play some amount of sub-optimal arms $\epsilon$ percent of the time for exploration purposes. 

\subsection{Adaptive Greedy (AG1) Strategy} \label{section:ag1}

In the context of delayed feedback the player must play $N$ arms before updating their belief about the system contained within each time window $\Omega$. In the the Adaptive Greedy (AG1) strategy, we opt to play the optimal arm based on the previous time epochs estimate of $\mu_t^k$ for $N\epsilon$ times, and leave ample room for exploration. We define a strategy to address the effect of non-stationarity in the reward function via an adaptive Bayesian method derived from the $\epsilon$-greedy strategy. In this scenario, the previous time epoch's parameter estimates are used to estimate the optimal arm $k_t^*$, at time $t$. We play the corresponding arm that maximizes the reward value based on the previous time window's estimate. For each time epoch containing N arms plays, we define $n_t^*$ as the number of times to play the optimal arm $k_t^*$. 

\begin{equation} \label{eq:eg_num_plays}
	n_t^* = \Bigl\lfloor N (1-\epsilon) \Bigr\rfloor\qquad
\end{equation}

And for non-optimal arms, 

\begin{equation} \label{eq:eg_num_plays_non_optimal}
	n_{t} = \Bigl\lceil N \frac{\epsilon}{K - 1} \Bigr\rceil\qquad
\end{equation}

The AG1 strategy ensures that parameter for estimates for $\mu_t^k$ are constantly being updated to reflect changes due to non-stationarity. We use the previous time frames observation window $\Omega_r$ to estimate $\mu_t^k$ for the current time epoch.

\begin{algorithm}[tbp]
\caption{AG1}\label{algo:ag1}
\begin{algorithmic}[1]
    \State {$Q = \emptyset$}
    \For {$t = 1 \to T$}
        \For {$k = 1 \to K$}
            \State {Compute $\widehat{\mu^k_t}$ from $\Omega_r$}
        \EndFor
        \State {$k = \operatorname*{argmax}_{k} \widehat{\mu^k_t} $}
        \For {$1 \to n_t^*$}
            \State {Play $k$}
            \State {$Q \gets Q(k) $}
        \EndFor
        \For {$k^{'} = 1 \to K$}
            \If {$k^{'} \neq k $}
                \For {$1 \to n_t$}
                    \State {Play $k^{'}$}
                    \State {$Q \gets Q(k^{'}) $}
                \EndFor
            \EndIf
        \EndFor
    \EndFor
\end{algorithmic}
\end{algorithm}

Similar to the $\epsilon$-greedy algorithm the AG1 strategy forces exploration of potentially sub-optimal arms. This gives rise to a linear lower bound on expected total regret. However, it does provide a means to detect new changes in the expected value of the reward function, as the process solely relies on the last time epoch's, $t-1$, estimate of the expected reward $\widehat{\mu^k_t}$.  The AG1 strategy is an algorithm that modifies the $\epsilon$-greedy's parameter estimation procedure such that it does not take into account historical value beyond that of the corresponding time window $\Omega_r$.

\subsection{UCB1 Algorithm}

A more robust solution to the MAB problem is the UCB1 algorithm introduced first in \citep{Lai:1985} and further analyzed in \citep{Auer:2002}. The UCB1 algorithm places a penalty on the number of times each arm is selected. The UCB1 algorithm can be applied in both the stationary and non-stationary case, with a logarithmic bound on regret, as evidenced in \citep{Garivier:2008}. UCB1 demonstrates a logarithmic bound on regret, as opposed to a constant bound imposed by $\epsilon$-greedy. However, the rate of convergence towards an optimal arm may be slower than other greedy strategies. We specify a UCB specific metric for the $k^{th}$ arm as $m_j$,

\begin{equation} \label{eq:ucbmetric}
    m_t^k = \mu^k_t + \sqrt{\frac{2 \log{t}}{ n(k) } }
\end{equation}

Where $n(k)$ refers to the number of time the $k^{th}$ arm was played within the observable time window. The action we take at time $k_t$ is expressed as the the arm that maximizes $m_t^k$,

\begin{equation} \label{eq:ucbcriteria}
    k_{t} = \operatorname*{argmax}_{k} m_t^k
\end{equation}

\begin{algorithm}
\caption{UCB1 Algorithm}\label{algo:ucb1}
\begin{algorithmic}[1]
    \State $Q = \emptyset$
    \For {$t = 0 \to T$}
        \For {$k = 1 \to K$}
            \State Compute $\widehat{\mu^k_t}$
        \EndFor
        \State Play $k_{t} = \operatorname*{argmax}_{k} m_t^k$
        \State $Q \gets Q(k) $  
    \EndFor
\end{algorithmic}
\end{algorithm}

The UCB1 algorithm imposes a logarithmic bound on the regret. For the stationary case this as been proven in the work of \citep{Auer:2002} expressed in  Eq. (\ref{eq:ucbregretbound}).

\begin{equation} \label{eq:ucbregretbound}
	\mathbb{E}[R_T(\pi)] \geq 8\sum^{}_{k: \mu_t^k < \mu_t^* }\Big(\frac{\log{n(k)} }{\mu_t^k -\mu_t^* } \Big) + \Big( 1 + \frac{\pi^2}{3}\Big)\Big( \sum^{K}_{k=1}\mu_t^k -\mu_t^* \Big)
\end{equation}

\subsection{Thompson Sampling}

\citep{Thompson:1933} introduced a Bayesian method framework for implementing MAB strategies which accounts for uncertain reward distributions. The advances in Bayesian methods have made it easier to apply randomized probability matching with almost any reward distribution \citep{Scott:2010}. \citep{Chapelle:2011} showed a highly competitive empirical result comparing Thompson Sampling with other MAB strategies. Thompson Sampling has some similarity with the $\epsilon$-greedy algorithm, the key difference being that $\epsilon$-greedy takes $\widehat{\mu^k_t}$ as the true estimate of the reward distribution and maximizes the expected reward with that assumption, whereas Thompson Sampling draws a random sample from an expected distribution parameters, $\theta$, playing the arm a with the probability that it is the optimal arm based on model parameters, $\widehat{\theta}$. The probability of selecting arm $k$, is expressed as, 

\begin{equation} \label{eq:thompson_integral}
	P(k) = \int \mathbb{I}\Big[ \mathbb{E}[Q(k)] = \max_{a} Q(k) \Big] P(\theta | R) d \theta
\end{equation}

Where the integral need not be computed explicitly, as it suffices to compute $\theta$ via a Bernoulli process, returning $\{0, 1\}$, as expressed in Eq. \ref{eq:bern_approx}. Each arm is played $\gamma$ * $n_t$ times, where $n_t$ is the number of stores assigned to that arm at time $t$, and that is the estimate of $\theta$ for that trial. Thus $\theta$ can be expressed as the grand average of the probability of success across all time epochs, expressed in Eq. \ref{eq:mu_calc}. We sample from the distribution $P(\hat{\theta}|Q)$, provided historical rewards $Q$.

\begin{equation} \label{eq:bern_approx}
	\hat{\theta}_k = \widehat{\mu_t^k} \in [0, 1]
\end{equation}

\begin{algorithm}[!h]
\caption{Thompson Sampling}\label{algo:Thomp}
\begin{algorithmic}[1]
    \State $Q = \emptyset$
    \For {$t = 0 \to T$}
        \State $\mathbf{Estimate}(\theta)$
        \For {$k \in \{1,..., K\}$}
            \State Compute $ \widehat{\mu_t^k} $
            \State Sample $ Q(k_t) \sim \widehat{\theta}_k$
        \EndFor
        \State $k_t = \operatorname*{argmax}_{k} Q(k_t) $
        \State $Q \gets Q(k_t)$ 
    \EndFor
\end{algorithmic}
\end{algorithm}

The Thompson Sampling strategy typically achieves superior performance in terms of regret minimization in comparison with other MAB strategies. We present evidence of this in Section \ref{subsec:stat-res}. A theoretical lower bound on regret is provided in \citep{Goyal:2012}, providing further detail. Serving as one of the few papers exploring the theoretical regret bounds on Thompson Sampling, \citep{Goyal:2012} proved that the lower bound on expected regret for Thompson Sampling strategies is logarithmic. In Section \ref{subsec:stat-res}, we will see empirical evidence of Thompson Sampling strategy being superior to other MAB strategies in the stationary reward scenario, in terms of regret minimization.

\section{Experimental Results}

In this section, we present the results of simulations pertaining to the cases of a stationary, and a non-stationary reward functions. The performance of the functions with respect to regret minimization is evaluated. In the stationary case, a fixed parameter $\mu^k$ is generated for each arm $k$. In the non-stationary case, a sinusoidal wave with respect a time is generated for $\mu^k_t$. 

\subsubsection{Stationary Scenario} \label{subsec:stat-res}

We compare the performance of three MAB strategies, $\epsilon$-greedy, UCB1, and Thompson Sampling in the stationary case. For $\epsilon$-greedy we set $\epsilon = 0.1$. A simulation environment similar to a live scenario is constructed. We provide $k$ arms, where $k = 10$. Subsequently, the MAB strategy will determine which algorithm is optimal. With observations up until the end of the last time epoch $t-1$. We specify 50 stores playing any of k arms over a period of 100 epochs.

\begin{figure}[!h]
  \includegraphics[width=\textwidth]{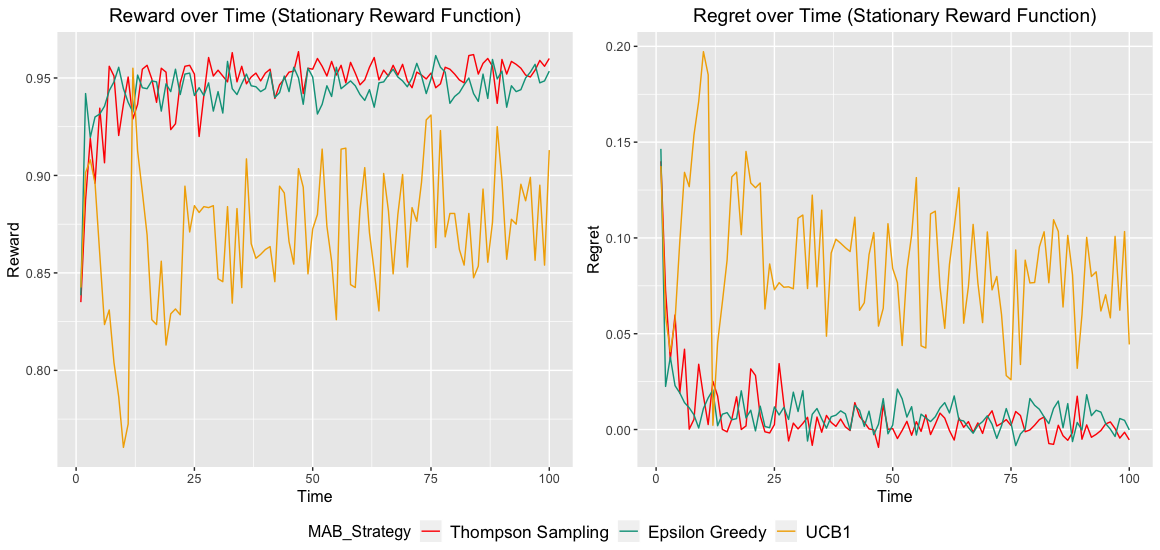}
  \caption{Regret over time for stationary reward function.}
  \label{fig:stat_plot}
\end{figure}

We present simulation results mimicking performance of the in store \textit{fill-rate} optimization algorithm for stationary reward distributions. In our simulation, the parameters are specified in Section \ref{subsec:delayed_feedback} where $N = 50$, $K = 10$, and $\gamma = 50$. This allows for a total of 2500 arm plays each epoch. We measure the results in Table \ref{table:stat_results}, on cumulative regret $R_T$ and cumulative reward $V_T$. Note that the regret for an epoch can be lower than 0 due to the stochastic nature of the simulated assortment reward possibly producing rewards higher than the highest probability assigned to an assortment algorithm. We examine that Thompson Sampling obtains the best results in terms of cumulative regret, however, the results are comparable to $\epsilon$-greedy when measured against the cumulative reward. The simulation results indicate that the strategy results in poor performance compared to it's counterpart strategies. We hypothesize this is due to the fact that the UCB1 strongly penalizes playing of the previous arm, causing it to over-explore less optimal strategies. Since in the delayed feedback scenario, arms must be played $\gamma$ times in order to obtain a reward measurement, this hinders the performance of UCB1. Suffice it is fair to acknowledge that the UCB1 strategy has a logarithmic regret bound that is not constantly bounded. However, the over-exploration of sub-optimal arms in the delayed feedback scenario make the UCB1 strategy a poor choice for our application. 

\begin{table}[h!]
\centering
\begin{tabular}{||c c c||} 
 \hline
 MAB Strategy & Cumulative Regret & Cumulative Reward \\ [0.5ex] 
 \hline\hline
 $\epsilon$-Greedy & 0.8928 & 94.45 \\ 
 Thompson Sampling & 0.7262 & 94.75 \\ 
 UCB1 & 8.7477 & 86.98 \\ [1ex] 
 \hline
\end{tabular}
\caption{Cumulative regret and rewards of various MAB strategies.}
\label{table:stat_results}
\end{table}

\subsubsection{Non-Stationary Scenario}

In this section we present the results of a MAB strategy for non-stationary reward functions. In one simulation we use two reward functions $K=2$, where the expected reward function $\mu^k_t$ is sinusoidal with respect to time. The sinusoids are not identical to one another, providing a time period where one strategy is strongly dominant over another strategy. During this period, the MAB strategy should determine which arm is optimal, and play that arm most frequently. We compare the AG1 strategy outlined in Section \ref{section:ag1} to the $\epsilon$-greedy algorithm, with a time window $\Omega_T = 3$ days. We observe that, in terms of regret minimization, the AG1 algorithm is superior to the $\epsilon$-greedy algorithm. The reasoning for this is due to the fact that the AG1 strategy retains the parameter estimates from the previous time window, without the need to re-estimate the parameter again.

\begin{figure}[!h] 
  \includegraphics[width=\textwidth]{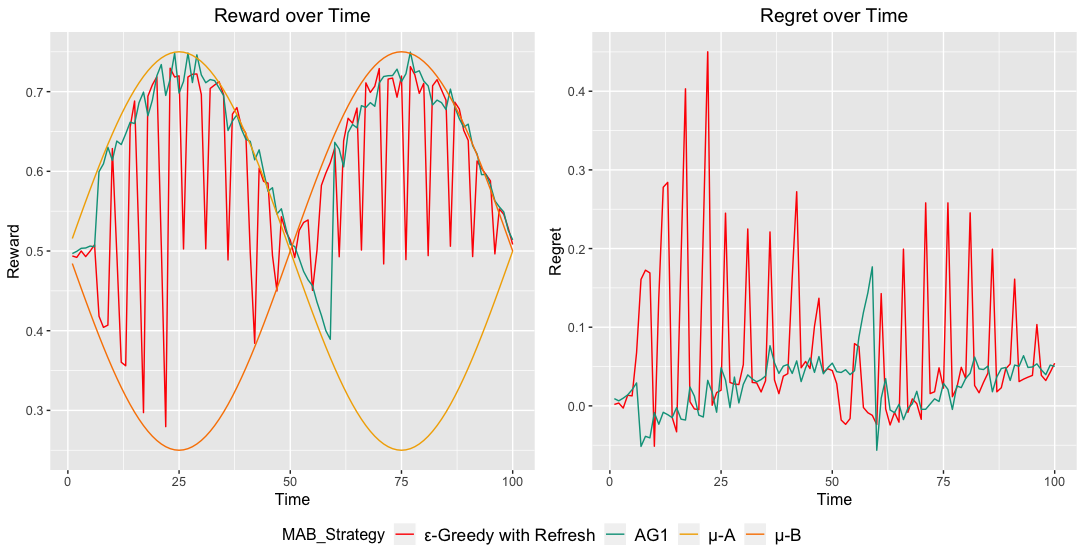}
  \caption{Regret over time for stationary reward function.}
  \label{fig:non_stat_plot}
\end{figure} 

As evidenced from Fig. \ref{fig:non_stat_plot} for illustrative purposes, we see that the AG1 algorithm is able to minimize the regret, by changing $k$ arm only when required to do so due to a regime change in the expected reward. The utilization of parameter memory, conducting a greedy strategy off of previous estimates of $\mu^k_{t-1}$ subsequently storing the data for future estimates provides a simple yet effective solution over an $\epsilon$-greedy that simply restarts at each time interval.

\begin{table}[h!]
\centering
\begin{tabular}{||c c c||} 
 \hline
 MAB Strategy & Cumulative Regret & Cumulative Reward \\ [0.5ex] 
 \hline\hline
 $\epsilon^*$-greedy & 6.881 & 58.92 \\ 
 AG1 & 2.784 & 63.08 \\ [1ex] 
 \hline
\end{tabular}
\caption{Simulation results using k = 2 arms (sinusoidal reward function).}
\label{table:stat_results_2_arm}
\end{table}

Subsequently, we investigate the results of this simulation for $K = 10$ arms. We include the results for $\epsilon^*$-greedy and $TS^*$, where * expresses the existence of a restart period. The restart period begins with no observations, playing each arm equally, until it re-discovers the optimal arm for the current time epoch, $t$. We see that $TS^*$ is able to perform better than $\epsilon^*$-greedy in terms of regret minimization, however, the AG1 algorithm performs significantly better for regret minimization due to its ability to remember the previous time window's, $\Omega_r$, and provide a parameter estimate based on those observations. 

\begin{table}[h!]
\centering
\begin{tabular}{||c c c||} 
 \hline
 MAB Strategy & Cumulative Regret & Cumulative Reward \\ [0.5ex] 
 \hline\hline
 $\epsilon^*$-greedy & 7.121 & 59.16 \\
 $TS^*$ & 6.241 & 59.82 \\ 
 AG1 & 2.558 & 63.71 \\ [1ex] 
 \hline
\end{tabular}
\caption{Simulation results using k = 10 arms (sinusoidal reward function).}
\label{table:stat_results_10_arm}
\end{table}

\section{Conclusion}

In the context of MAB strategies, we have observed the efficacy of multiple strategies in the face of delayed feedback in the stationary and non-stationary cases for reward functions. In the stationary case, Thompson Sampling displays the most promising results in terms of cumulative total regret. The UCB1 displayed poor results for our non-stationary use case, however we acknowledge that there exists adaptive tuning methods for UCB1, that can be further explored to minimize regret. We hypothesize that this is due to the delayed feedback mechanism reducing exploration too strongly after some time epochs have passed, subsequently causing the strategy to select non-optimal arms. In the non-stationary case, we devise an MAB strategy based on $\epsilon$-greedy. It performs significantly better than either Thompson Sampling or $\epsilon^*$-greedy. The Bayesian approach to yields significantly better performance as the model parameters are estimated in the prior time window, and retained for the next time epoch estimation. 

For any number of arms, the AG1 strategy is shown to be effective at accounting for the non-stationarity of the reward function. In the case where a simulated sinusoidal wave was used to generate the reward function, regret was significantly minimized as the AG1 strategy suffered less cumulative regret that other MAB strategies with a restart period.  In future work, it would be useful to compare the effectiveness of our model against other adaptive $\epsilon$-greedy strategies, such as the one proposed in \citep{Tokic:2011}. Furthermore, it would be useful to consider tuning methods that can reduce the cumulative regret for cases regarding UCB1. The results of this simulation provide sound evidence that the AG1 provides superior performance for regret minimization in the MAB framework specifically in the scenario of delayed feedback, and non-stationary reward processes. Further areas of improvement can include a customization of Thompson Sampling to include a means to apply the Adaptive strategy for parameter estimation. We recognize the potential for MAB strategies to have a drastic impact in the optimization of experimentation for many industry application, and due to the realization that non-stationary reward functions do occur, we conducted this simulation to obtain further insight into such processes.

\vskip 0.2in
\bibliography{sample}

\end{document}